\documentclass{article}

\usepackage[preprint]{neurips_2024}

\usepackage[utf8]{inputenc} 
\usepackage[T1]{fontenc}    
\usepackage{hyperref}       
\usepackage{url}           
\usepackage{booktabs}       
\usepackage{amsfonts}      
\usepackage{nicefrac}       
\usepackage{microtype}     
\usepackage{xcolor}      
\usepackage{multirow}         
\usepackage{graphicx}
\usepackage{subfigure}
\usepackage{makecell}
\usepackage{amssymb}
\usepackage{amsmath}
\usepackage{algorithm}
\usepackage{algorithmic}
\usepackage{ulem}
\usepackage{wrapfig}
\usepackage{multicol}
\usepackage{amsthm}
\usepackage{flushend}
\usepackage{enumerate,enumitem}
\usepackage{caption,subcaption}
\usepackage{booktabs}

\title{Improving Data-aware and Parameter-aware Robustness for Continual Learning}

\author{
  Hanxi Xiao$^1$\quad Fan Lyu$^{2}$\thanks{Corresponding author.}\\
  $^1$ Department of Information Management, University of Shanghai for Science and Technology\\
$^2$ New Laboratory of Pattern Recognition, Institute of Automation, Chinese Academy of Sciences\\
  \texttt{hanxi.xiao@st.usst.edu.cn}, \texttt{fan.lyu@cripac.ia.ac.cn} \\
}

\begin{document}

\maketitle

\begin{abstract}
  The goal of Continual Learning (CL) task is to continuously learn multiple new tasks sequentially while achieving a balance between the plasticity and stability of new and old knowledge. 
  This paper analyzes that this insufficiency arises from the ineffective handling of outliers, leading to abnormal gradients and unexpected model updates. To address this issue, we enhance the data-aware and parameter-aware robustness of CL, proposing a \textbf{R}obust \textbf{C}ontinual \textbf{L}earning (RCL) method. 
  From the data perspective, we develop a contrastive loss based on the concepts of uniformity and alignment, forming a feature distribution that is more applicable to outliers. From the parameter perspective, we present a forward strategy for worst-case perturbation and apply robust gradient projection to the parameters. 
  The experimental results on three benchmarks show that the proposed method effectively maintains robustness and achieves new state-of-the-art (SOTA) results. The code is available at: \url{https://github.com/HanxiXiao/RCL}
\end{abstract}

\section{Introduction}

Continual learning (CL) refers to learning a series of different tasks in sequence without forgetting previous knowledge~\cite{wang2024comprehensive}.
The major challenge of CL is called catastrophic forgetting, which means that when a model is trained on a new task, its performance on previously learned tasks decreases. 
To alleviate catastrophic forgetting, numerous methods are proposed and utilized in CL, such as replay-based~\cite{chaudhry2019tiny,riemer2018learning,shin2017continual,lyu2021multi,sun2022exploring} and optimization-based~\cite{saha2021gradient,yang2023data,deng2021flattening} methods.
Most of these methods are pursuing the improvement of plasticity and stability. Plasticity refers to the ability of the model to learn new knowledge, and stability refers to the ability of the model to maintain old knowledge. 

In addition to plasticity and stability, \textit{robustness} is also important for CL, that is, model training should not be easily affected by unexpected input data~\cite{croce2020reliable}.  
Recent research on improving the robustness of CL mainly focus on how to combine with classic adversarial training methods, and can be divided into two categories. 
First, combining adversarial training with task learning \cite{chou2022continual,ghosh2021adversarial}, but adding adversarial examples to the training set increases learning difficulty and leads to catastrophic forgetting. Second, the memory-replay method stores data that is beneficial for robustness~\cite{mi2023adversarial,wang2022improving}, but this requires additional storage and the stored data does not contain any outlier samples.
To the best of our knowledge, the causes of insufficient robustness in CL is under-explored.

\begin{figure}
\centering
\includegraphics[width=.9\textwidth]{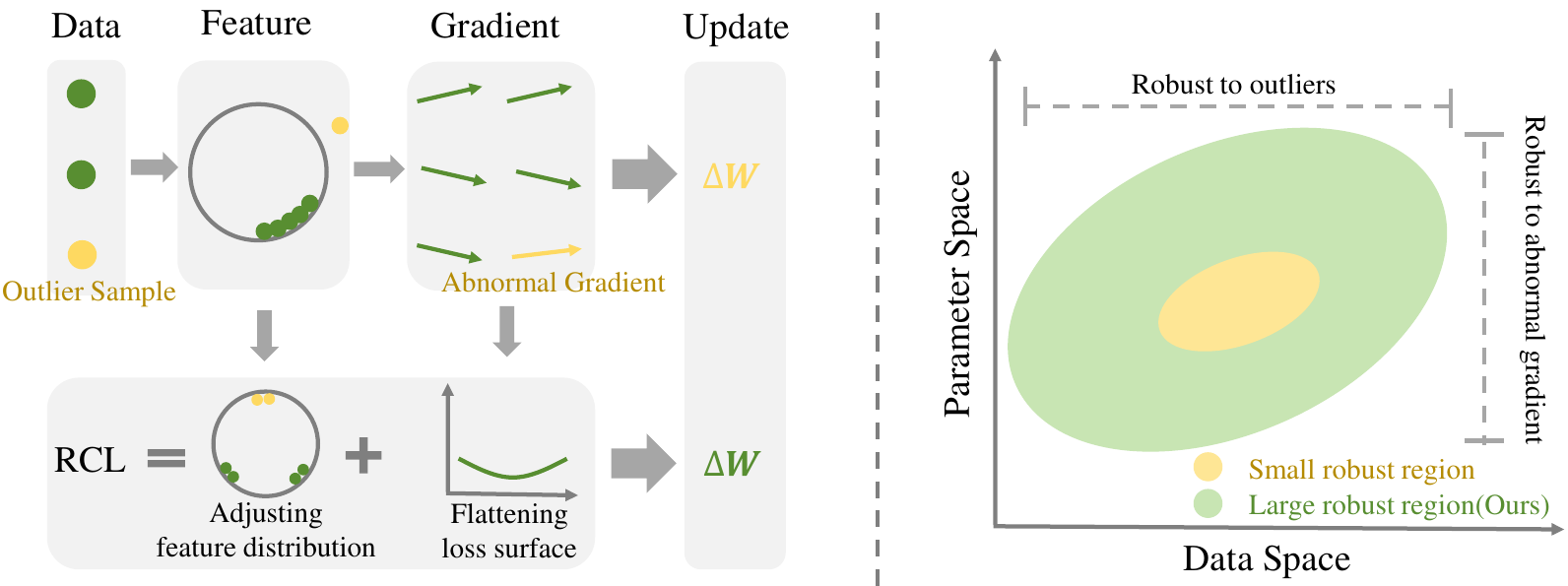}
\caption{Outlier samples can generate abnormal gradients that hinder continual learning. Our method reduces the impact from both data and parameter perspectives (Left). Our method has a wider robust region, and even as the model continuously learns new tasks and the impact of abnormal gradients increases, our method can still remain undisturbed (Right).}
\vspace{-10px}
\label{fig1}
\end{figure}

In this paper, we study the robustness of CL from data and parameter perspectives. 
We first analyze the impact of outlier samples on the CL model and find that the abnormal gradients generated by outlier samples not only hinder the learning of the current task, but also accelerate the forgetting of past tasks. Further analysis of abnormal gradients reveals that improving the weight loss landscape of the CL method can effectively alleviate the impact of abnormal gradients, yielding small robust region, as shown in Fig. \ref{fig1}.

Based on the analysis of the generation of outlier samples and abnormal gradients, we propose a novel \textbf{R}obust \textbf{C}ontinual \textbf{L}earning (RCL) method to reduce the impact of abnormal gradients from both data and parameter dimensions.
First, we optimize the feature distribution to reduce class collapse phenomenon, making the feature distribution uniform and aligned, so that it is not easy to confuse with other class features when facing outlier samples. 
Second, for the update of parameter, we improve the flatness of the loss surface by adding random and worst-case perturbations, thereby alleviating the impact of abnormal gradients on CL tasks under different weight spans. 
The final gradient obtained is called a robust gradient. 
At the same time, Gradient Projection Memory (GPM)~\cite{saha2021gradient} restricts the back-propagation robust gradient to the orthogonal direction with the key gradient subspace of the previous task, so that after adjusting the weights to adapt to the new task, it can also ensure consistent prediction results of the previous task data. 
Our contributions are three-fold:

\begin{enumerate}[label=(\arabic*),left=0pt,itemsep=0pt]
    \vspace{-5px}
    \item We analyze the robustness of continual learning from a gradient perspective and demonstrate how outlier samples affect the performance of continual learning.
    \item In data-aware robustness, we introduce uniformity and alignment loss to optimize the distribution of features on the unit hypersphere, helping to reduce the impact of outlier samples.
    \item In parameter-aware robustness, we introduce random perturbations to extend uniformity into the parameter space and alleviate the impact of abnormal gradients by optimizing flatness in the worst-case perturbation direction.
\end{enumerate}

\section{Related Work}
\label{relate}

\textbf{Continual learning.} CL methods can be divided into five categories based on their main motivations and implementation forms \cite{wang2024comprehensive}. 
1) \textit{Regularization-based Methods.} These methods \cite{zenke2017continual,cha2020cpr, du2023multi} balance old and new tasks by adding explicit regularization terms \cite{kirkpatrick2017overcoming} and calculating the importance of network weights \cite{aljundi2018memory} for each old task to prevent significant changes in the important parameters of old tasks. 2) \textit{Replay-based Methods.} These methods improve model performance by storing old training samples \cite{chaudhry2019tiny,riemer2018learning,lyu2023measuring,liu2023centroid} and training additional generative models \cite{shin2017continual,wu2018memory} to replay the generated data or features during training. 3) \textit{Architecture-based Methods.} These methods \cite{ahn2019uncertainty,yoon2017lifelong,mallya2018packnet,serra2018overcoming} dynamically adjust the model size and add a new set of learnable parameters for each new task. 4) \textit{Representation-based Methods.} These methods create and leverage representation advantages, combined with self-supervised learning (SSL) \cite{pham2021dualnet,madaan2021representational,gallardo2021self} and large-scale pre-training \cite{wu2019large,mehta2023empirical,shi2022mimicking}. 5) \textit{Optimization-based Methods.} A typical operation among these methods is to perform gradient projection. GPM \cite{saha2021gradient} completes the gradient projection of new tasks by storing the gradient subspace of old tasks in memory. Some studies \cite{deng2021flattening,liang2023adaptive,yang2023data} indicate that GPM based methods have potential for enhancing model performance. 
In this work, we also chose GPM as the baseline. 

\textbf{Robustness of Continual Learning.}
Deep neural networks can effectively solve many difficult machine learning tasks. However, existing neural networks are easily vulnerable to abnormal data such as attacks \cite{carlini2017towards}. 
In the field of continual learning, strategies for improving model robustness can be roughly divided into two categories: 1) \textit{Adversarial training.} \cite{chou2022continual}  integrates continual learning and progressive adversarial training to improve both corruption robustness and structure compactness of the defensive model. \cite{mi2023adversarial} adds adversarial samples of new and old tasks to the training dataset by generating them. 
2) \textit{Memory-based method.} \cite{mi2023adversarial} reducing gradient confusion by storing robust and diverse past data, thereby reducing the negative impact of adversarial training and robust perceptual experience replay. \cite{wang2022improving} improve robustness by storing the coreset \cite{jia2023robustnesspreserving} related to robustness in memory. In addition, some works \cite{wu2021pretrained,yoon2019scalable,farquhar2018towards} focus on the robustness of continual learning under different experimental conditions, such as task order, memory constraints, computational constraints, or time constraints, while others focus on robust continual learning models against backdoor attacks \cite{umer2020targeted} or privacy protection \cite{hassanpour2022differential}. However, unlike the aforementioned studies, we do not directly combine adversarial training with continual learning, nor do we store excess data in memory. We conduct a thorough analysis of the impact of outlier data on the robustness of continual learning.

\section{Relating Robustness of Continual Learning to Gradient}
\label{analysis}

\subsection{Problem Formulation and Robust Gradient}

Robustness has been proven to be crucial for neural networks~\cite{mangal2019robustness}, as insufficient robustness can lead to model performance being easily affected. 
In CL, due to data availability and model size constraints, robustness becomes even more critical during dynamic sequential learning processes.  
In this paper, we claim that insufficiently robust gradients contribute to the lack of robustness in CL models.

To illustrate this, we first give the definition of the CL task.
Given $T$ sequential tasks, CL receives each task with a dataset $\mathcal{D}=\{(x_{i}, y_{i})\}_{i=1}^{N}$, where $x_{i}$ and $y_{i}$ represent the input data point and the corresponding label. 
All CL tasks share a backbone that builds on a neural network $f(\mathbf{W}, \cdot)$. 
The goal of CL is to achieve effectiveness on all seen tasks in the sequence learning.
However, CL can only observe samples of the current task, making CL challenging. 
The expression for minimizing the loss of the current task $t$ is as follows:
\begin{equation}
    \label{eq1}
    \mathop{\min}_{\mathbf{W}}\mathbb{E}_{(x_{i}^{t}, y_{i}^{t})\sim\mathcal{D}^{t}}\left[\mathcal{L}(f(\mathbf{W},x_{i}^{t}),y_{i}^{t})\right],
\end{equation}
where $\mathcal{L}(\cdot,\cdot)$ is the loss function such as cross-entropy. 
The model is updated by stochastic gradient descent and the gradient $\mathbf{g} ={\partial\mathcal{L}(\mathbf{W}, X)}/{\partial\mathbf{W}}$.
For a data point $x$, if it is an outlier or an adversarial example, the CL model will consequently acquire abnormal gradients.
Regarding model parameters $\mathbf{W}$, if the parameters are situated at a local optimum for a certain task, introducing new tasks or anomalous data will degrade the performance of the parameters for that task.
In the following, we will introduce how do the two factors influence the gradient, i.e., the model parameter $\mathbf{W}$ and the data input $X$, and make a CL model underrobust.

\subsection{Outlier Samples Causing Abnormal Gradient}
\label{3.2}
When there is an abnormality in the dataset, we can represent it by adding perturbation $\epsilon$ to a normal data point $x$. When the model learns outlier samples, the calculation of its loss function follows a first-order Taylor expansion as follows:
\begin{equation}
        \mathcal{L}(\mathbf{W},x+\epsilon)=\mathcal{L}(\mathbf{W},x)+\nabla_{x}\mathcal{L}^{T}\epsilon,\quad \text{s.t.} \Vert\epsilon\Vert_p \le \rho.
        \label{eq:outlier}
\end{equation}
From Eq.~\eqref{eq:outlier}, the outlier sample generates the loss an excess term $\nabla_{x}\mathcal{L}^{T}\epsilon$, which is the gradient along the perturbation direction. When the perturbation is large, the generation of this abnormal gradient causes the loss value to increase, resulting in misclassification. 

In CL, let the output of the data before softmax be $r(x)$, i.e. $f(x)=\text{softmax}(r(x))$ and assume the model has learned $t-1$ tasks, each containing multiple categories. 
When learning task $t$, $r_{p}(x_{t})$ means the logits of past classes of $x_t$, $r_{c}(x_{t})$ means the logits of current classes of $x_t$. $p$ represents the category that has already been learned, and $c$ represents the category that is currently being learned. When generating outlier samples $\Tilde{x}_t=x_t+\epsilon$ for the task $t$ being learned, $\Tilde{x}_t$ raises the logits corresponding to other categories. According to \cite{madry2017towards}, so there are: $r_{p}(\Tilde{x}_t)>r_{p}(x_{t}),\ r_{c}(\Tilde{x}_t)<r_{c}(x_{t})$. 
Due to the gradient amplification property of the outlier samples, we choose the cross-entropy as the loss function, and the abnormal gradients of the current data to the past task are increased:
\begin{equation}
    \label{neg}
        \frac{\partial\mathcal{L}}{\partial r_p(\Tilde{x}_t)}=f_p(\Tilde{x}_t)-y_p>f_p(x_t)-y_p=\frac{\partial\mathcal{L}}{\partial r_p(x_t)},
\end{equation}
where $y_p=0$ is the label of $x_{t}$ on past tasks. Eq.~\eqref{neg} means the abnormal gradient generated by outlier samples not only affects the learning of current data, but also amplifies the gradient of current data on past tasks, accelerating forgetting. 

\subsection{The Potential Impact of Abnormal Gradients on Flatness}
\label{3.3}
In this section, we try to explore the impact of increasing abnormal gradients. 
In a single-layer network structure, there is $r(\Tilde{x}_t)=\sigma({W}x_{t}+b)$, and it is worth noting that the derivation still holds below the multi-layer network structure and convolutional layers (see Appendix \ref{con}). 
According to the chain differentiation rule, the gradient can be represented by 
\begin{equation}
        \frac{\partial \mathcal{L}}{\partial \mathbf{W}}=\frac{\partial \mathcal{L}}{\partial r_p(\Tilde{x}_t)}\cdot\frac{\partial r_p(\Tilde{x}_t)}{\partial \mathbf{W}}=\left(\frac{\partial \mathcal{L}}{\partial r_p(\Tilde{x}_t)} \odot \sigma \right)\Tilde{x}_t.
\end{equation} 
In the following, we show that when the outlier samples generate abnormal gradients about the previous task, it also increases the span of parameter updates.

\begin{figure}
  \centering
  \includegraphics[width=1\textwidth]{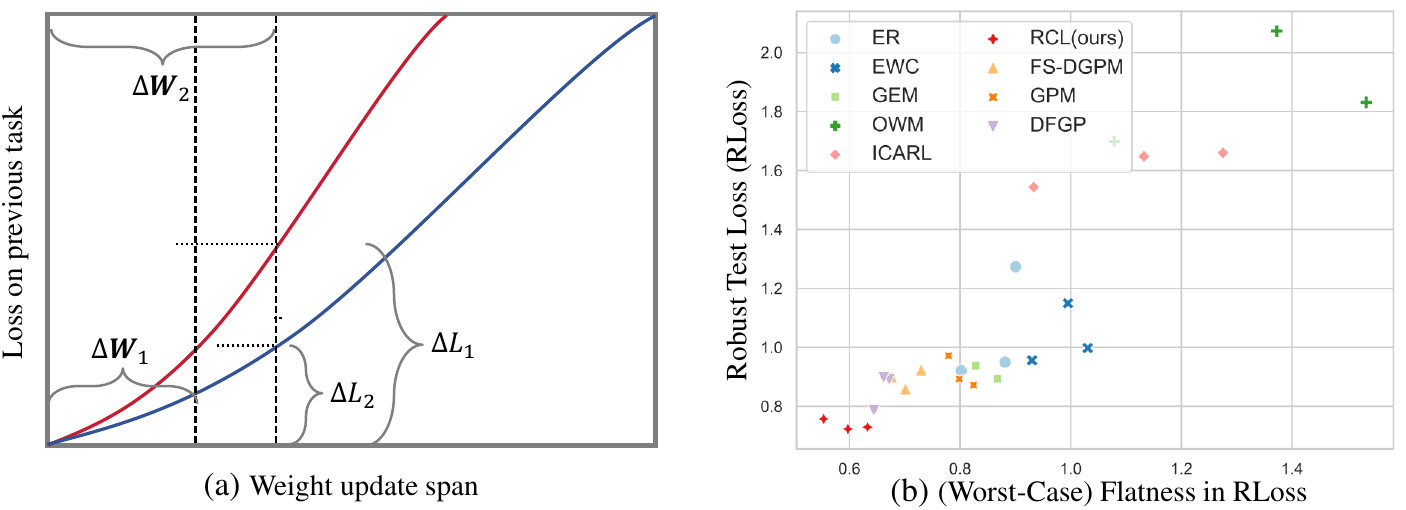} 
  \caption{(a) The impact of weight span on the loss value of previous tasks. (b) The relationship between robust loss and flatness.}
  \label{flatness}
  \vspace{-10px}
\end{figure}

As shown in Fig. \ref{flatness}(a), when the weight span increases from $\Delta \mathbf{W}_{1}$ to $\Delta \mathbf{W}_{2}$, the loss value of the previous task becomes higher, indicating an intensification of forgetting about the previous task. If we do not consider the changes in weight span and instead compare the differences in loss surfaces with different levels of flatness, we find that a flatter loss curve (blue-line) has less variation in loss values than an uneven loss curve (red-line) under the same span. From this, it can be seen that abnormal gradients reduce the performance and robustness of the model during the learning process, and a flatter loss surface can better alleviate this phenomenon. Intuitively, a flatter loss surface can be more stable and robust in the face of both enlarged and reduced weight spans. 

The relationship between robustness and flatness of some adversarial training methods was validated in \cite{stutz2021relating}, and we extend it to CL task. 
We denote the loss of the model on adversarial samples as robust loss, i.e., cross-entropy loss on FGSM ($\mu=0.001$) adversarial examples \cite{madry2017towards}. 
In the field of CL, FS-DGPM \cite{deng2021flattening} selects some random directions when calculating flatness, but we choose adversarial weight directions and we define the worst-case flatness as:
\begin{equation}
    \label{worst-case}
        \mathop{\max}_{\upsilon}\left[\mathcal{L}(f(\mathbf{W}_{t}+\xi \cdot \upsilon,x_{i}^{t}),y_{i}^{t})\right]-\mathcal{L}(f(\mathbf{W}_{t},x_{i}^{t}),y_{i}^{t}).
\end{equation}
To eliminate the scaling invariance of DNNs, we follow \cite{wu2020adversarial,deng2021flattening} and perform perlayer normalization: $\upsilon^{(l)}=\frac{\upsilon^{(l)}}{\Vert \upsilon^{(l)} \Vert_{2}}\Vert \omega^{(l)} \Vert_{2}, \upsilon \in \mathbb{R}^{\mathbf{W}}$ and $\xi$ is a very small value representing the magnitude of weight $\omega$ movement. As shown in Fig. \ref{flatness} (b), we select eight different CL methods, each with three experiments with different parameter settings. It can be seen that under the same weight span, the flatter the loss surface, the smaller the robust loss. 

To summarize the analysis, outlier samples can generate abnormal gradients that lead to misclassification. During the training process, these outlier samples can generate abnormal gradients about the previous task, altering the weight distribution. This alteration can negatively impact the learning of subsequent tasks. A flat weight loss landscape can mitigate this issue by reducing the impact of weight perturbation changes on the model's robustness.

\section{Method}
\label{method}
\begin{figure}[t]
  \centering
  \includegraphics[width=.9\textwidth]{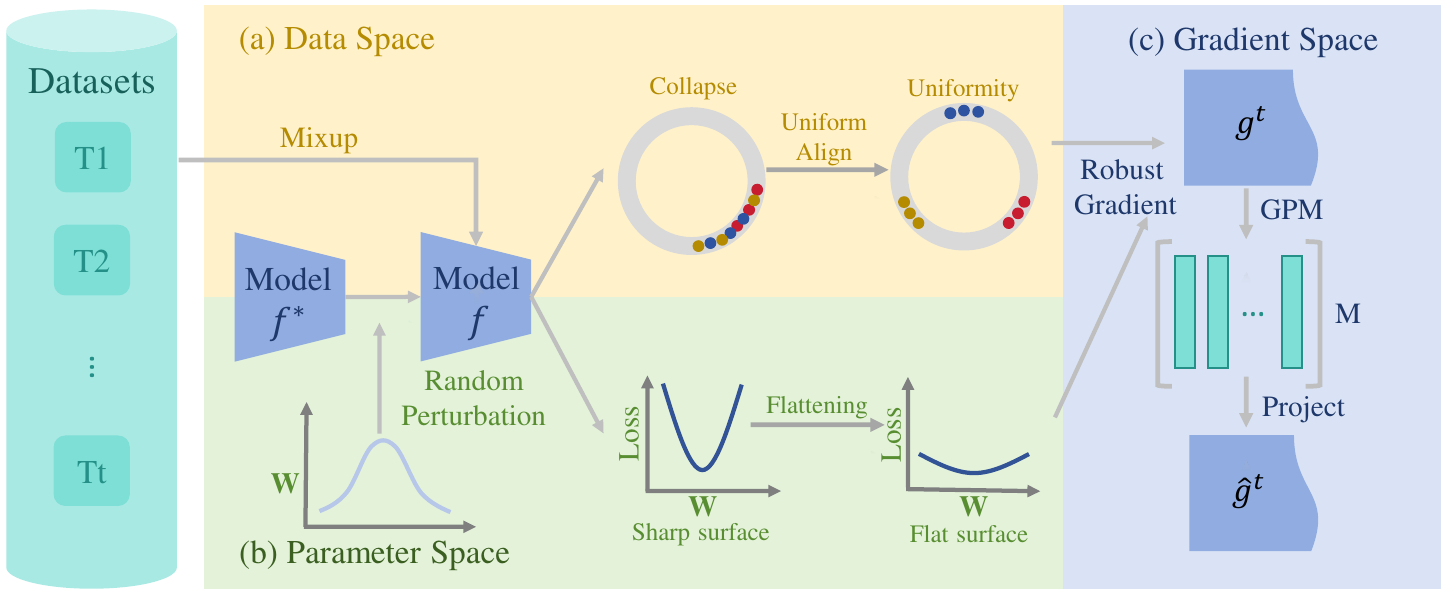}
  \caption{An overview of RCL. (a) Data space. Distribute features uniform and aligned on the unit hypersphere. (b) Parameter space. Added random perturbations and worst-case perturbations to flatten the loss surface. (c) Gradient space. Using GPM to complete gradient projection.}
  \label{view}
\vspace{-10px}
\end{figure}

In this section, we provide RCL method as shown in Fig. \ref{view}.
In Sec. \ref{data}, we mitigate the impact of outlier samples by adjusting the distribution of data features. In Sec. \ref{para}, we reduce the impact of abnormal gradients by adding random and worst-case perturbations. Then introduce the overall algorithm in Sec. \ref{overall}. 

\subsection{Robustness of Feature Distribution across Sequential Tasks}
\label{data}

First, we propose to mitigate the impact of outlier samples, thereby preventing the generation of abnormal gradients during the CL training process. 
We cannot determine if the input data to the model are outliers in advance, but we can approach this from the perspective of data feature distribution. By making the distribution of outlier samples as close as possible to that of normal samples, we can achieve more typical gradients.

Inspired by~\cite{xu2018spherical,mettes2019hyperspherical}, we assume the feature space is an unit hypersphere. 
The feature distribution sensitive to outlier samples is shown in Fig. \ref{view} (a) collapse distribution, different category features are mapped to the same position on the hyper sphere after being processed by the model, which makes it difficult for the model to distinguish between outlier samples and normal samples, resulting in misclassification. Therefore, we consider a uniform and aligned distribution of data features on a hypersphere, as shown in the Fig. \ref{view} (a) uniformity distribution. Specifically, uniformity ensures that features cover the entire sphere while preserving as much feature information as possible. Alignment brings the same class closer together and further separates different classes, which can make decision boundaries clearer. In this way, when encountering outlier samples, the model can find the most similar features to them.

In order to meet the requirements of CL, $p_\mathrm{t}$ is the data distribution over $\mathbb{R}^{n}$ and $t \in \{1,\dots,T\}$. 
Similar to the work in \cite{wang2020understanding}, we consider the Gaussian potential kernel (a.k.a. Radial Basis Function kernel) and define the uniformity loss as the logarithm of the average pairwise Gaussian potential:
\begin{equation}
    \begin{split}
    \label{unif}
        \mathcal{L}_\mathrm{uniform}(f;\tau;t)&\triangleq \mathrm{log}\ \mathbb{E}_{x,y\stackrel{\mathrm{i.i.d}} \sim p_\mathrm{t}}\left[e^{-\tau\Vert f(x)-f(y) \Vert_{2}^{2}}\right],\quad \tau>0,
    \end{split}
\end{equation}
The alignment loss is straightforwardly defined with the expected distance between positive pairs: 
\begin{equation}
    \label{align}
        \mathcal{L}_\mathrm{align}(f;\alpha)\triangleq-\mathbb{E}_{(x,y)\sim p_\mathrm{pos}}\left[\Vert f(x)-f(y) \Vert_{2}^{\alpha}\right],\quad \alpha>0,
\end{equation}
$\tau$ and $\alpha$ are fixed parameters. Note that when we optimize the feature distribution within a single task, as the number of CL tasks increases, our feature space also continual to expand. 

We generate sample pairs through data augmentation, and then obtain data-aware robust gradient by differentiating uniformity and alignment loss.
Specifically, randomly select any two samples $(x_{i}^{t},y_{i}^{t})$ and $(x_{j}^{t},y_{j}^{t})$ in a minibatch from task $t$, and then use vanilla Mixup to generate new sample $(\Tilde{x}^{t},\Tilde{y}^{t})$: $\Tilde{x}^{t}=\gamma x_{i}^{t}+(1-\gamma)x_{j}^{t}, \Tilde{y}^{t}=\gamma y_{i}^{t}+(1-\gamma)y_{j}^{t}$, where $\gamma \sim$ Beta$(\alpha,\alpha)$ and $\gamma \in $ [0,1]. Afterwards, two pairs of positive samples can be obtained, namely $(x_{i}^{t},\Tilde{x}^{t})$ and $(x_{j}^{t},\Tilde{x}^{t})$.

Inspired by \cite{yang2023data}, we perform a further optimization on $\gamma$ whose goal is to maximize the loss corresponding to the samples after augmentation. We optimize the overall distribution by optimizing the worst positive pair distribution. Specifically, we will add a perturbation $\varepsilon$ to the sampled $\gamma$ in the neighborhood of radius $\rho$, so that the loss on the new image mixed with perturbation is the worst. The formal expression of gamma’s (i.e., $\Tilde{\gamma}=\gamma+\varepsilon$) goal is as follows:
\begin{equation}
    \label{gamma}
    \mathop{\max}_{\Vert\varepsilon\Vert_2\le\rho}\mathcal{L}^{t}(f(\mathbf{W},\Tilde{x}^{t}),\Tilde{y}^{t})+\mathcal{L}_\mathrm{ua}^{t}(x_{i}^{t},x_{j}^{t},\Tilde{x}^{t}).
\end{equation}
However, it is difficult to exactly find the solution of Eq. \eqref{gamma}, therefore, the optimization objective is transformed into a first-order Taylor approximation problem to solve the worst-case data perturbation $\hat{\varepsilon}$ (see Appendix~\ref{solu}). We not only optimize the cross-entropy loss of the new image $(\Tilde{\gamma}=\gamma+\hat{\varepsilon})$ and the original image, but also optimize the uniformity and alignment loss: 
\begin{equation}
    \begin{aligned}
    \label{lua}
        \mathcal{L}_\mathrm{ua}^{t}(x_{i}^{t},x_{j}^{t},\Tilde{x}^{t})=\mathrm{log}\ \mathbb{E}\left[e^{-t\Vert f(x_{i}^{t})-f(x_{j}^{t}) \Vert_{2}^{2}}\right]
        +\frac{1}{2}\left(\mathbb{E}\left[\Vert f(x_{i}^{t})-f(\Tilde{x}^{t}) \Vert_{2}^{\alpha}\right]
        +\mathbb{E}\left[\Vert f(x_{j}^{t})-f(\Tilde{x}^{t}) \Vert_{2}^{\alpha}\right]\right).
    \end{aligned}
\end{equation}
Eq. \eqref{lua} will also be involved in solving the worst-case parameter perturbations and gradients.

\subsection{Robustness of Parameter by Worst-case Perturbation}
\label{para}
In Sec \ref{3.3}, we have linked the robustness of CL to flatness and aim to improve robustness by flattening the loss surface.
To achieve parameter-aware robustness, we design a two-step optimiation approach.
First, we optimize the parameter distribution by adding random perturbations to the parameters and utilizing uniformity loss. 
Second, we use adversarial weight perturbations from \cite{deng2021flattening,yang2023data}, which can help the model find a flatter loss surface, then, obtain parameter-aware robust gradient.
  
As indicated by \cite{dinh2017sharp,Ramasinghe2023HowMD}, flatness is highly sensitive to reparametrization and initialized parameters. Therefore, before learning the current task, we add random perturbations $\mathbf{\phi}$ to the parameters to help weight updates find a flatter loss landscape in the future.
For each $\mathbf{\theta}$ in the model parameter $\mathbf{W}$, use the reparameterization trick as follows:
\begin{equation}
    \label{eq6}
        \mathbf{\Tilde{\theta}} = \mathbf{\theta} + \epsilon \cdot \mathbf{\phi},\quad \epsilon \sim \mathcal{N}(0,1),
\end{equation}
where $\epsilon$ is the value randomly sampled from a Gaussian distribution, and $\mathbf{\phi}$ is the trainable parameter whose value range needs to be adjusted accordingly for different models.
The distribution of parameters in high-dimensional space should be similar to a uniform distribution, so we also choose the uniformity loss as the target:
\begin{equation}
    \label{phi}
    \mathop{\min}_{\mathbf{\phi}}\mathcal{L}_\mathrm{uniform}=\mathrm{log}\ \mathbb{E}_{\mathbf{\theta},\mathbf{\Tilde{\theta}}\stackrel{\mathrm{i.i.d}} \sim p_\mathrm{w}}\left[e^{-\tau\Vert \mathbf{\theta}-\mathbf{\Tilde{\theta}} \Vert_{2}^{2}}\right],\quad \tau>0,
\end{equation}
where $\mathbf{\Tilde{\theta}}$ will replace $\mathbf{\theta}$ as the initialization parameter of the model to complete the following training. 
 
However, the addition of random perturbations does not guarantee direct optimization of the flatness of the loss surface.
Therefore, during training, we consider optimizing the flatness in the \textit{worst-case direction}, i.e., the direction with the fastest gradient descent.
Specifically, the objective of the weight perturbation is to find the $\upsilon$ that causes the worst-case loss, i.e.,
\begin{equation}
    \label{wc}
        \mathop{\max}_{\Vert \upsilon \Vert_{2} \le \rho}\mathcal{L}^{t}(f(\mathbf{W}+\upsilon,x^{t}),y^{t})+\lambda\mathcal{L}^{t}(f(\mathbf{W}+\upsilon,\Tilde{x}^{t}),\Tilde{y}^{t})+\kappa\mathcal{L}_\mathrm{ua}^{t}(x_{i}^{t},x_{j}^{t},\Tilde{x}^{t}).
\end{equation}
Performing weight perturbations on augmented data is to allow weight perturbations to be explored in a wider space. The addition of uniformity and alignment loss also ensures that the distribution of data in space is taken into account when searching for worst-case weight perturbations. 
$\rho$ represents the domain radius for finding the perturbation $\upsilon$ for $\mathbf{W}$. By solving the first-order Taylor approximation problem of Eq. \eqref{wc}, the solution of worst-case perturbation $\hat{\upsilon}$ can be obtained as follows: 
\begin{equation}
    \label{ups}
        \hat{\upsilon} \approx \rho \cdot \frac{\nabla_{\mathbf{W}}\mathcal{L}^{t}(f(\mathbf{W},x^{t}),y^{t})+\lambda\mathcal{L}^{t}(f(\mathbf{W},\Tilde{x}^{t}),\Tilde{y}^{t})+\kappa\mathcal{L}_\mathrm{ua}^{t}(x_{i}^{t},x_{j}^{t},\Tilde{x}^{t})}
        {\Vert \nabla_{\mathbf{W}}\mathcal{L}^{t}(f(\mathbf{W},x^{t}),y^{t})+\lambda\mathcal{L}^{t}(f(\mathbf{W},\Tilde{x}^{t}),\Tilde{y}^{t})+\kappa\mathcal{L}_\mathrm{ua}^{t}(x_{i}^{t},x_{j}^{t},\Tilde{x}^{t}) \Vert_{2}},
\end{equation}
where $\lambda$ and $\kappa$ are hyperparameters that can be used to adjust the proportion of loss. After obtaining perturbation $\hat{\upsilon}$, we can obtain new weights $\hat{\mathbf{W}}\ (\hat{\mathbf{W}}=\mathbf{W}+\hat{\upsilon})$. Then, we can simply minimize the loss value under the worst-case parameter $\hat{\mathbf{W}}$ to improve flatness.

\subsection{Robust Gradient Projection}
\label{overall}

Although the obtained gradients have shown significant improvement in robustness, in order to further alleviate the problem of catastrophic forgetting, we perform gradient projection \cite{saha2021gradient} operations on the robust gradient. 
Based on the description in the first two sections, we obtain the robust gradient that is robust from both data and parameter perspectives:
\begin{equation}
    \label{og}
        g^{t} \approx \nabla\left(\mathcal{L}^{t}(f(\hat{\mathbf{W}},x^{t}),y^{t})+\lambda\mathcal{L}^{t}(f(\hat{\mathbf{W}},\Tilde{x}^{t}),\Tilde{y}^{t})+\kappa\mathcal{L}_\mathrm{ua}^{t}(x_{i}^{t},x_{j}^{t},\Tilde{x}^{t})\right).
\end{equation}
We calculate the core gradient space to construct the projection matrix $\mathbf{M}$. 
We first randomly select $n$ samples to obtain the representation $\mathbf{R}$ of each layer $l\in\{1,\cdots,L\}$, and perform Singular Value Decomposition (SVD) on this representation $\mathbf{R} = \mathbf{U\Sigma V}$, to select the top-$k$ most important basis. Then, we use the eigenvectors corresponding to the top-$k$ maximum eigenvalues as the core gradient space to form the projection matrix $\mathbf{M}\ (\mathbf{M}=\{\mathbf{M}_{l,t}\}_{l=1}^{L})$. 
We obtain the projected gradient: 
\begin{equation}
    \hat{g}^{t}=g^{t}-\mathbf{M}(\mathbf{M})^{\top}g^{t}.
\end{equation}
Then, the model is updated with the $\hat{g}^{t}$, that is: $\mathbf{W}^l=\mathbf{W}^l-\eta\cdot\hat{g}^{t}$.  
The detailed GPM process is provided in the Appendix \ref{gpm}.  
The overall algorithm can be found in the Appendix \ref{rcl}.

\begin{table}
\caption{Results of different CL methods on three datasets. MTL uses the entire dataset to learn all tasks simultaneously in a single network, it can be used as an upper bound for CL learning.}
    \label{acc}
    \centering
    \resizebox{\linewidth}{!}{
    \renewcommand{\arraystretch}{1.3}
    \begin{tabular}{lccccccccc}
    \toprule
    \multicolumn{2}{l}{\multirow{2}*{Method}}&\multicolumn{2}{c}{CIFAR-100 (10 Tasks)}&&\multicolumn{2}{c}{5-Datasets (5 Tasks)}&&\multicolumn{2}{c}{MiniImageNet (20 Tasks)}\\
    \cmidrule(r){3-4}\cmidrule(r){6-7}\cmidrule(r){9-10}
    && ACC(\%) & BWT(\%)&&ACC(\%) & BWT(\%)&&ACC(\%) & BWT(\%)\\
    \hline
    MTL&&80.67$\pm$0.42&-&&93.15$\pm$0.16&-&&84.16$\pm$1.26&-\\
    \hline
    EWC \cite{kirkpatrick2017overcoming}&&68.77$\pm$0.76&-2.17$\pm$1.03&&88.90$\pm$0.26&-4.35$\pm$1.32&&52.01$\pm$2.02&-12.52$\pm$3.39\\
    ER \cite{chaudhry2019tiny}&&71.83$\pm$0.66&-6.54$\pm$0.89&&89.01$\pm$0.31&-3.69$\pm$0.49&&58.94$\pm$0.85&-7.16$\pm$0.77\\
    A-GEM \cite{chaudhry2018efficient}&&63.75$\pm$1.96&-15.14$\pm$2.43&&84.02$\pm$0.28&-4.06$\pm$0.11&&56.98$\pm$0.88&-7.24$\pm$1.32\\
    OWM \cite{zeng2019continual}&&51.74$\pm$0.36&-1.03$\pm$0.41&&-&-&&-&-\\
    TRGP \cite{lin2022trgp}&&73.39$\pm$0.28&-0.28$\pm$0.15&&90.94$\pm$0.11&-0.13$\pm$0.02&&62.52$\pm$0.74&-0.23$\pm$0.42\\
    GPM \cite{saha2021gradient}&&73.01$\pm$0.36&-0.12$\pm$0.23&&90.02$\pm$0.00&-2.15$\pm$0.37&&61.61$\pm$3.02&-0.34$\pm$0.31\\
    FS-DGPM \cite{deng2021flattening}&&73.95$\pm$0.27&-2.61$\pm$0.22&&-&-&&61.03$\pm$0.75&-1.28$\pm$0.18\\
    API \cite{liang2023adaptive}&&74.29$\pm$0.58&-0.27$\pm$0.32&&90.70$\pm$0.40&-0.30$\pm$0.20&&65.91$\pm$0.60&-0.33$\pm$0.18\\
    DFGP \cite{yang2023data}&&74.12$\pm$0.30&-0.85$\pm$0.42&&91.56$\pm$0.17&-1.62$\pm$0.04&&69.88$\pm$0.79&-0.10$\pm$0.55\\
    \hline
    RCL&&\textbf{75.43$\pm$0.09}&-0.07$\pm$0.00&&\textbf{92.00$\pm$0.16}&-1.59$\pm$0.19&&\textbf{71.34$\pm$0.38}&1.53$\pm$0.23\\
    \bottomrule
    \end{tabular}
    }
    \vspace{-15px}
\end{table}

\section{Experiments}

\subsection{Experimental Setup}
\textbf{Datasets:} We evaluate CL methods on three widely used datasets, including Split CIFAR-100 \cite{krizhevsky2009learning}, Split MiniImagenet, and 5-Datasets \cite{ebrahimi2020adversarial}. The CIFAR-100 Split is constructed by randomly dividing 100 classes of CIFAR-100 into 10 tasks with 10 classes per task. The Mini-Imagenet split is constructed by splitting 100 classes of Mini-Imagenet into 20 tasks, and each task consists of 5 classes. 5-Datasets is a continual learning benchmark with 5 different datasets, including CIFAR10, MNIST, SVHN, notMNIST, and Fashion-MNIST. 

\textbf{Baselines:}
We compare RCL against multiple methods. 
Multitask Learning (MTL); 1) Memory-based: ER \cite{chaudhry2019tiny}, A-GEM \cite{chaudhry2018efficient} 2) Regularization-based: EWC  \cite{kirkpatrick2017overcoming} 3) Orthogonal-Projection-based: OWM \cite{zeng2019continual}, TRGP \cite{lin2022trgp}, GPM \cite{saha2021gradient}; 4) GPM-based: FS-DGPM \cite{deng2021flattening}, API \cite{liang2023adaptive}, DFGP \cite{yang2023data}.

\textbf{Metrics:} 
Following existing works, we use average final accuracy (ACC) and backward transfer (BWT) as evaluation metrics. ACC is the average accuracy of all tasks. BWT measures forgetting. The formulas of these two metrics are computed by:
        $ACC=\frac{1}{T} \sum\nolimits_{i=1}^{T} A_{T,i}$ and $BWT=\frac{1}{T-1} \sum\nolimits_{i=1}^{T-1} A_{T,i}-A_{i,i}$,
where $T$ is the number of tasks, $A_{t,i}$ is the test classification accuracy of the model on $i$-th task after learning the last sample from $t$-th task.

\subsection{Experimental Result}

\textbf{Performance:} In Tab. \ref{acc}, the accuracy of RCL is significantly improved over previous work on all datasets, with achieving the best average accuracy 75.43$\%$, 92.00$\%$, 71.34$\%$. 
Our RCL achieves the accuracy gains of 1.14$\%$, 0.44$\%$ and 1.46$\%$ on the three datasets CIFAR-100, 5-Datasets and MiniImageNet, respectively, compared to the best baseline methods API and DFGP.
In terms of accuracy gain, MiniImageNet (20 Tasks) > CIFAR-100 (10 Tasks) > 5-Datasets (5 Tasks) indicates that our method exhibits better performance in solving difficult problems. 
This is because old tasks are more likely to be forgotten when the number of tasks increases.
In these experiments, we observe that our method has also made great progress in reducing forgetting. For example, our method reduces forgetting by 0.78$\%$, 0.03$\%$ and 1.63$\%$ compared to DFGP on three datasets, respectively. It is worth noting that the forgetting rate on the CIFAR-100 dataset is only -0.07$\%$, while a positive number of 1.53$\%$ appears on the MiniImageNet dataset, which is much higher than other baseline methods.

\begin{figure}
\begin{minipage}[b]{0.45\textwidth}
    \centering
    \includegraphics[width=.9\linewidth]{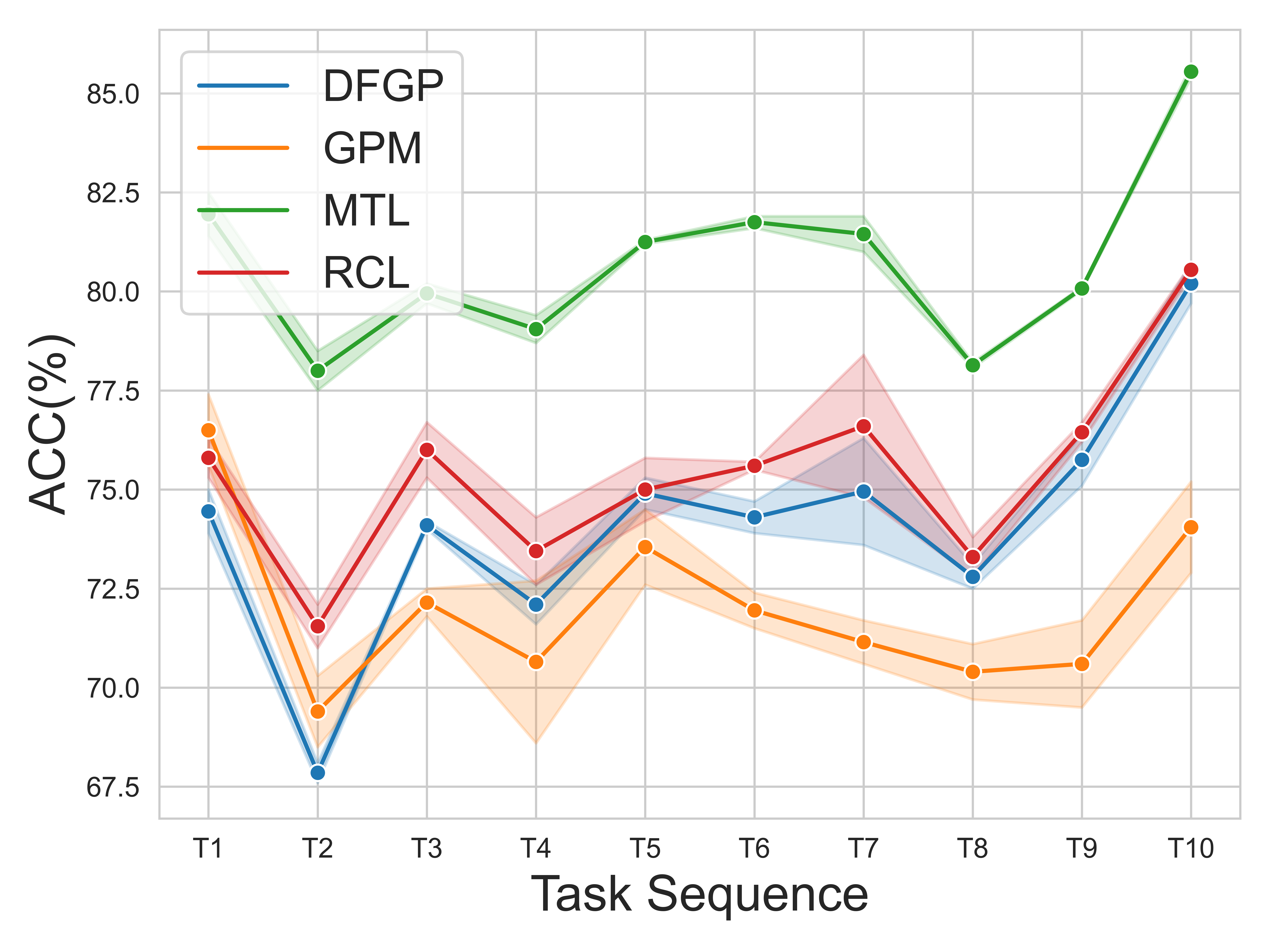}
    \caption{New task accuracy on CIFAR-100.}
    \label{task}
\end{minipage}
~~~~
\begin{minipage}[b]{0.45\textwidth}
    \centering
    \setlength{\tabcolsep}{4pt}
    \renewcommand{\arraystretch}{1.5}
    \captionof{table}{Ablation study on CIFAR-100 and MiniImageNet.}
    \scalebox{0.8}{
    \begin{tabular}{ccccccc}
        \toprule
        \multicolumn{2}{c}{\multirow{2}*{Method}}&\multicolumn{2}{c}{CIFAR-100}& &
        \multicolumn{2}{c}{MiniImageNet} \\
        \cmidrule(r){3-4}\cmidrule(r){6-7}
        & & ACC(\%) & BWT(\%) & & ACC(\%) & BWT(\%)\\
        \hline
        DFGP & &74.12 & -0.85 & & 69.88 & -0.10\\
        \hline
        w/W($\phi$) & &74.54 & -0.83 & & 70.06 & 0.60\\
        w/D(ua) & &75.16 & -0.60 & & 70.50 & 0.82\\
        w/G(ua) & &75.26 & -0.66 & & 70.02 & 0.60\\
        \hline
        RCL & & 75.43 & -0.07 & & 71.34 & 1.53\\
        \bottomrule
    \end{tabular}}
    \label{ablation}
\end{minipage}
\vspace{-10px}
\end{figure}

\textbf{Stability and plasticity analysis.} As shown in Fig. \ref{task}, we plot the performance of GPM, DFGP and RCL on the same line graph for new tasks. We observe that when learning more difficult tasks such as T2, T4, and T8, our method has an average improvement of 2.15$\%$, 1.35$\%$, and 0.5$\%$ compared to GPM and DFGP. Overall, although our method still lags behind MTL in performance on new tasks, it has improved compared to other baselines methods. This demonstrates that we can improve plasticity by accurately improving model robustness. As shown in Fig. \ref{flatacc} (c) and (d), we plotted the performance of GPM and RCL in ten tasks of CIFAR-100. For example, when task T10 is learned, the accuracy of GPM on tasks T2, T4, and T6 is 67.9$\%$, 69.5$\%$, and 72.4$\%$, respectively. RCL is 72.1$\%$, 72.6$\%$, 75.5$\%$, respectively. These evidences suggest that our method improves both robustness and continual learning performance, i.e. the stability and plasticity. 

\subsection{Ablation Study}

\textbf{Effectiveness of each component.} 
Our method adds two new components compared to DFGP, including random perturbations in the parameters (w/W($\phi$)) and uniformity and alignment loss (w/D(ua)) in the data distribution. As shown in Tab. \ref{ablation}, the accuracy of the model decreased when we only performed one robustness, with a significant decrease when we only performed (w/W($\phi$)), proving the effectiveness of each component. For example, when we only use (w/W($\phi$)) and (w/D(ua)) on CIFAR-100, the ACC decreases from 75.43$\%$ to 74.54$\%$ and 75.16$\%$, respectively. Furthermore, when searching for data perturbations $\hat{\varepsilon}$ and weight perturbations $\hat{\upsilon}$, we did not include alignment and uniformity loss (see details in Appendix \ref{rcl}), but only added it when obtaining gradients (w/G(ua)). We found that its ACC also decreased from 75.43$\%$ to 75.26$\%$, which also indicates that the addition of $\mathcal{L}_\mathrm{ua}^{t}$ can help us find more beneficial perturbations for model learning.

\begin{figure}[t]
\centering
    \begin{minipage}{0.24\linewidth}
		\centering
		\includegraphics[width=1\linewidth]{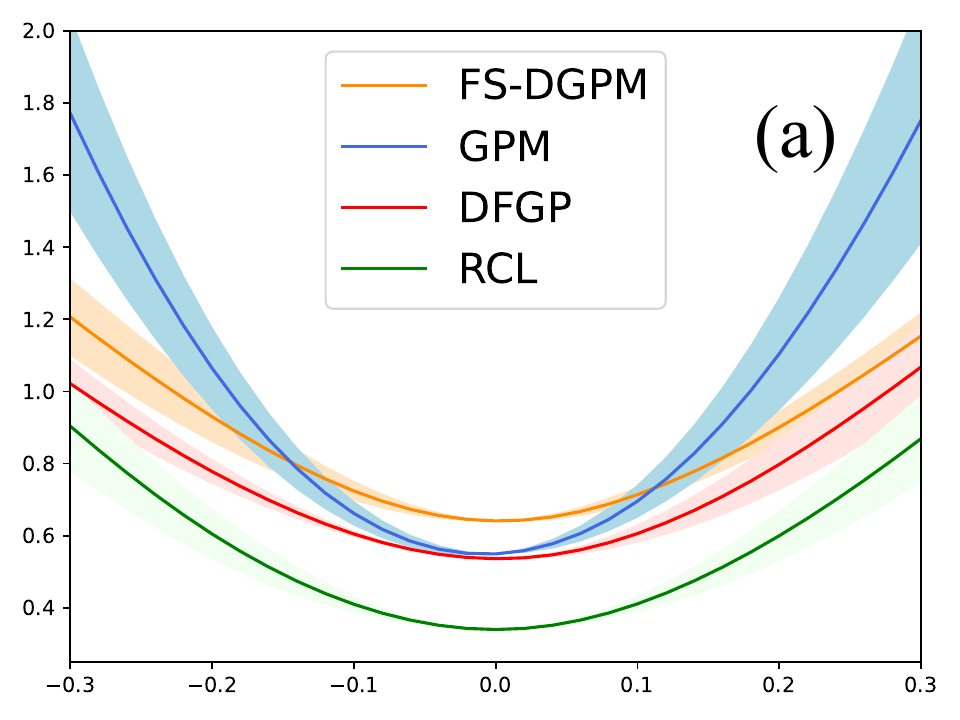}
	\end{minipage}
	\begin{minipage}{0.24\linewidth}
		\centering
		\includegraphics[width=1\linewidth]{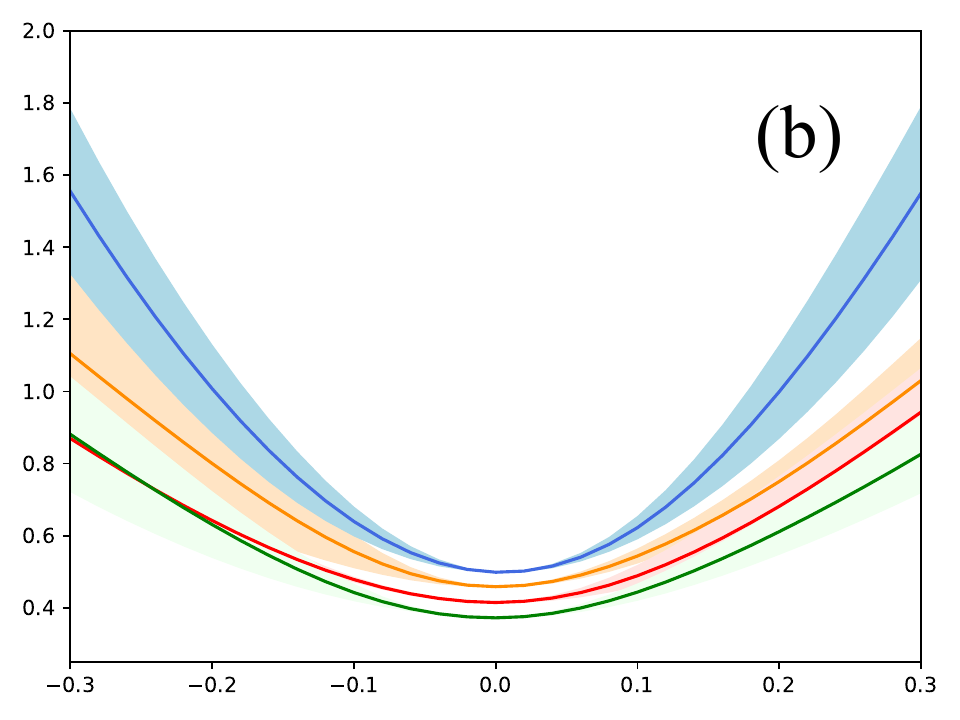}
	\end{minipage}
	\begin{minipage}{0.24\linewidth}
		\centering
		\includegraphics[width=1\linewidth]{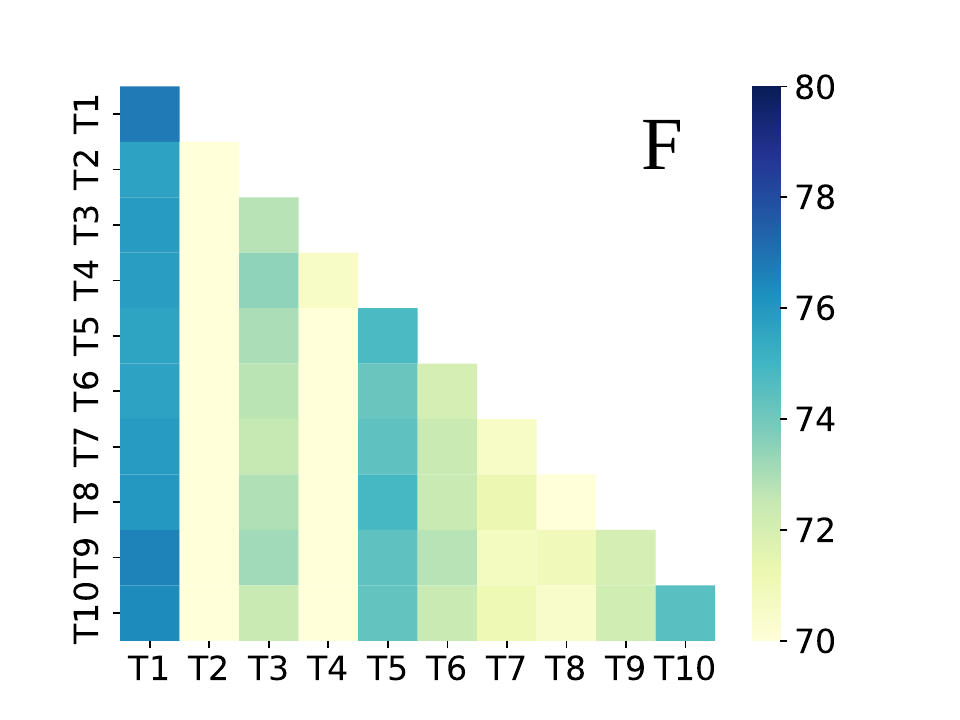}
	\end{minipage}
	\begin{minipage}{0.24\linewidth}
		\centering
		\includegraphics[width=1\linewidth]{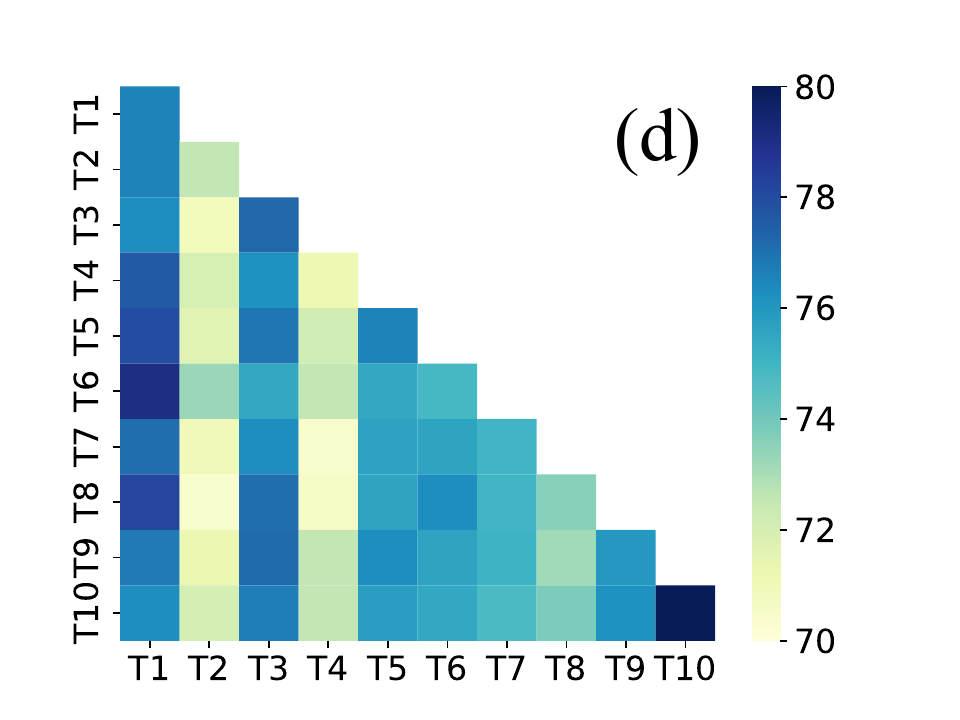}
	\end{minipage}
    \caption{The visualization of flatness and accuracy on CIFAR-100. (a) is the weight loss landscape of the second task after learning all ten tasks; (b) is the weight loss landscape of the fifth task after learning five tasks; (c) ACC of GPM; (d) ACC of RCL (darker is better).}
    \label{flatacc}
    \vspace{-10px}
\end{figure}

\textbf{Flatness visualization.} In order to more intuitively demonstrate the improvement of our method on overall flatness, we use the visualization method for flatness in FS-DGPM and randomly selected ten directions to draw an average case weight loss landscape. As shown in Fig. \ref{flatacc}, it can be observed that our method has the lowest loss value and the flattest loss landscape. 

\begin{wrapfigure}[16]{r}{7cm}
\vspace{-15px}
\centering
\includegraphics[width=0.50\textwidth]{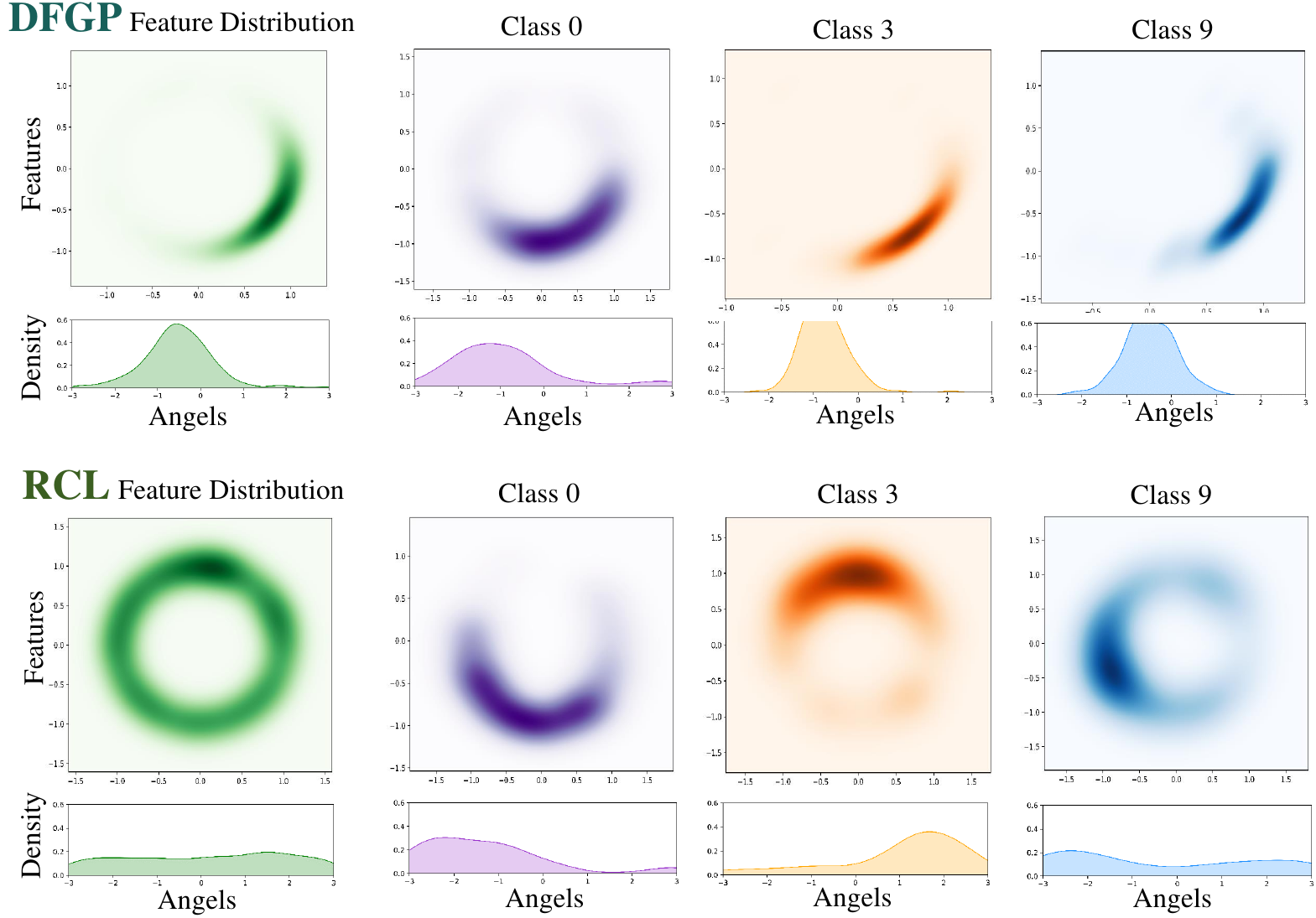}
  \caption{The visualization of uniformity on CIFAR-100. We plot feature distributions with Gaussian kernel density estimation (KDE) in $\mathbb{R}^2$ and KDE on angles (i.e., $\arctan 2(y, x)$ for each point $(x, y) \in S^{1}$). Three rightmost plots visualize feature distributions of selected specific classes. }
  \label{unit}
\end{wrapfigure}

\textbf{Uniformity visualization.} As shown in Fig. \ref{unit}, we randomly select 200 samples from each of the ten tasks in CIFAR-100, input them into the network after training, and adjust the output to 2-dimension to obtain the distribution of features on the unit hypersphere. Our method distributes features uniformly and separates different categories (see Fig. \ref{unit} below), allowing the model to preserve as much data information as possible and reduce overlap between categories. 
However, without the addition of alignment and uniformity loss, all feature information in DFGP is mapped to similar positions (see Fig. \ref{unit} upper), which is not conducive to the model learning new tasks and reducing forgetting.

\textbf{Robustness analysis.} To verify that our method indeed improves the robustness of the model, we used FGSM to generate adversarial perturbations of the input image, with the following generation rules: $X^{adv}=X+\mu \cdot sign(\nabla_{\mathbf{X}}(\mathcal{L}^{t}(f(\mathbf{W},X),Y))$, where $\mu$ is a hyperparameter of adversarial strength, and $X$ represents the samples of the test set. As shown in Tab. \ref{FGSM}, as $\mu$ increases, the accuracy of all methods decreases, but our method has the least decrease. For example, when $\mu$ is 0.01, on MiniImageNet dataset, GPM and DFGP decreased by 35.83$\%$ and 38.3$\%$ respectively, while our method only decreased by 12.75$\%$; when $\mu$ is 0.0001, on CIFAR-100 dataset, the first two have decreased 0.57$\%$ and 0.65$\%$, while our method remains unchanged. 
\begin{table}[t]
    \caption{Averaged accuracy inputting generated adversarial samples.}
    \label{FGSM}
    \centering
    \resizebox{.9\linewidth}{!}{
    \renewcommand{\arraystretch}{1.3}
    \begin{tabular}{ccccccc}
    \toprule
    &\multicolumn{3}{c}{CIFAR-100}&\multicolumn{3}{c}{MiniImageNet}\\
    \cmidrule(r){2-4}\cmidrule(r){5-7}
    $\mu$&GPM&DFGP&RCL&GPM&DFGP&RCL\\
    \hline
    0.0 &73.01(-0.00)&74.12(-0.00)&75.43(-0.00)&61.61(-0.00)&69.88(-0.00)&71.34(-0.00)\\
    \hline
    1e-04&72.44(-0.57)& 73.47(-0.65)&75.43\textbf{(-0.00)}&61.16(-0.45)&69.51(-0.37)&71.06\textbf{(-0.28)}\\
    1e-03&71.44(-1.57) & 72.27(-1.85)&74.01\textbf{(-1.42)}&56.59(-5.02)&64.95(-4.93)&69.52\textbf{(-1.82)}\\
    1e-02&58.60(-14.41)& 59.18(-14.94)&62.01\textbf{(-13.42)}&25.78(-35.83)&31.58(-38.3)&58.59\textbf{(-12.75)}\\
    \bottomrule
    \end{tabular}
    }
    \vspace{-10px}
\end{table}

\section{Conclusion and Limitation}
\label{lim}
In this paper, we analyzed the impact of outlier samples on CL performance and explained the impact of loss surface flatness on robustness from the perspective of abnormal gradients. Then, we proposed the RCL, which consists of two parts: robustness of feature distribution and robustness of parameter. The core idea is to adjust the distribution of features and find a flatter loss landscape. A large number of experiments show that our proposed RCL not only improves the robustness of CL, but also enhances the stability of old tasks and the plasticity of new tasks. 
Finally, we visualized the feature distribution and loss landscape, and it is evident that our method not only improves the uniformity and alignment of feature distribution, but also the flatness of the weight loss landscape. 
However, our method still has some \textit{limitations}, our method is not effective in online manner. In the future, we plan to further improve the robustness of our method in online setting.

\normalem
% \bibliographystyle{plain}
% \bibliography{ref}

\newpage
\appendix

\begin{Large}
    \centerline{\textbf{Appendix of Robust Continual Learning}}
\end{Large}

\section{Proof and Pseudo-code}

\subsection{Linear Transformation in Convolutional Layers}
\label{con}
For the convolution layer, We denote single-layer forward propagation as $r(\Tilde{x}_{t})=\sigma(\mathbf{W}\ast x_{t}+b)$, where $\mathbf{W} \in \mathbb{R}^{C_{O}\times C_{I}\times k\times k},\ x_{t} \in \mathbb{R}^{C_{I}\times h_{I}\times w_{I}},$ and $r(\Tilde{x}_{t})\in \mathbb{R}^{C_{O}\times h_{O}\times w_{O}}$. $C_{I}$ and $C_{O}$ denote input and output channels, respectively. $k$ denotes kernel size. $(h_{I}, w_{I})$ and $(h_{O}, w_{O})$ denote the input and output size of the feature, respectively. Through the reshaping process, reshape $\mathbf{W}$ into $(C_{I}\times k\times k)\times C_{O}$, $x_{t}$ into $(C_{I}\times k\times k)\times (h_{O}, w_{O})$, $r(\Tilde{x}_{t})$ into $C_{O} \times (h_{O}, w_{O})$. Then the convolution operation $\ast$ can be transformed into matrix multiplication, that is:
\begin{equation}
    \hat{r}(\Tilde{x}_{t})=\sigma(\hat{\mathbf{W}}\hat{x}_{t}+b).
\end{equation}
After that, the conclusion in Eq. \eqref{eq:outlier} about linear layer can be applied to the convolution layer. 

\subsection{The Solution of Worst-case Data Perturbation}
\label{solu}
We can obtain the solution for $\hat{\varepsilon}$ in $\Tilde{\gamma}$ by solving the first-order Taylor approximation problem of the maximization equation $ \mathcal{L}^{t}(f(\mathbf{W},\Tilde{x}^{t}),\Tilde{y}^{t})+\mathcal{L}_\mathrm{ua}^{t}(x_{i}^{t},x_{j}^{t},\Tilde{x}^{t})$, $\Tilde{\gamma}=\gamma+\varepsilon$ \cite{yang2023data}, and the solution $\hat{\varepsilon}$ is:
\begin{equation}
\begin{split}
    \label{var}
        \hat{\varepsilon} &\approx \mathop{\arg\max}\limits_{\Vert \varepsilon \Vert_2\le \rho}\mathcal{L}^{t}(f(\mathbf{W},\Tilde{x}^{t}),\Tilde{y}^{t})+\kappa\mathcal{L}_\mathrm{ua}^{t}(x_{i}^{t},x_{j}^{t},\Tilde{x}^{t})\\
        &+\varepsilon^{T}\left(\nabla_{\gamma}\mathcal{L}^{t}(f(\mathbf{W},\Tilde{x}^{t}),\Tilde{y}^{t})+\kappa\nabla_{\gamma}\mathcal{L}_\mathrm{ua}^{t}(x_{i}^{t},x_{j}^{t},\Tilde{x}^{t})\right)\\
        &=\rho \cdot \frac{\nabla_{\gamma}\mathcal{L}^{t}(f(\mathbf{W},\Tilde{x}^{t}),\Tilde{y}^{t})+\kappa\nabla_{\gamma}\mathcal{L}_\mathrm{ua}^{t}(x_{i}^{t},x_{j}^{t},\Tilde{x}^{t})}{\Vert \nabla_{\gamma}\mathcal{L}^{t}(f(\mathbf{W},\Tilde{x}^{t}),\Tilde{y}^{t})+\kappa\nabla_{\gamma}\mathcal{L}_\mathrm{ua}^{t}(x_{i}^{t},x_{j}^{t},\Tilde{x}^{t}) \Vert_{2}},
\end{split}
\end{equation}
the derivation process of Eq. \eqref{ups} is similar to it. Note that, observing the forms of the two optimization problems of Eq. \eqref{var} and Eq. \eqref{ups}. We can find that the optimal weight perturbation $\hat{\varepsilon}$ and data perturbation $\hat{\upsilon}$ can be computed together through a single backpropagation.

\subsection{Pseudo-code for Updating GPM}
\label{gpm}
As shown in Alg. \ref{alg2} for updating GPM, we first randomly sample $B_{n}^{t}$ from the training set to construct the representation matrix $R^{l}$ for each layer. Next, we perform SVD on $\mathbf{R}^{l}=U^{l}\Sigma^{l}(V^{l})^{\top}$ to select the top-$k$ most important basis according to the following criteria for the given threshold $\epsilon_{th}$(see line 10 in Alg. \ref{alg2}): 
\begin{equation}
    \label{eq14}
        \Vert \mathbf{M}_{l}(\mathbf{M}_{l})^{\top}\mathbf{R}^{l} \Vert_{F}^{2}+\Vert \hat{R}_{k}^{l} \Vert_{F}^{2} \ge \epsilon_{th} \Vert R^{l} \Vert_{F}^{2}
\end{equation}
where, $\Vert \cdot \Vert_{F}$ is the Frobenius norm of the matrix and $\epsilon_{th}$ ($0<\epsilon_{th}\ge 1$) is the threshold hyperparameter for layer $l$.
\begin{algorithm}[H]
    \renewcommand{\algorithmicrequire}{\textbf{Input:}}
	\caption{Algorithm of UpdateGPM} 
	\label{alg2} 
	\begin{algorithmic}[1]
		\REQUIRE bases matrix $\mathbf{M}$, network $f$ with $L$-layer, threshold $\epsilon_{th}$ for each layer, training set $\mathcal{D}_{train}^{t}$.
        \STATE$B_{n}^{t} \gets (X^{t},Y^{t}) \sim \mathcal{D}_{train}^{t}$
        \STATE$\mathcal{R} \gets $forward$(B_{n}^{t},f),\  $where$\ \mathcal{R}=\{(R^{l})_{l=1}^{L}\}$ \textcolor{blue}{\COMMENT{construct representation matrix by forward pass.}}
        \FOR{layer $l=1,2,...,L$} 
        \IF{$t>1$}
            \STATE $\hat{R}^{l}=R^{l}-\mathbf{M}_{l}(\mathbf{M}_{l})^{\top}R^{l}$ \textcolor{blue}{\COMMENT{remove duplicate representations.}}
            \ELSE
            \STATE $\hat{R}^{l}=R^{l}$
            \ENDIF
            \STATE$U^{l} \gets$ SVD$(\hat{R}^{l})$ \textcolor{blue}{\COMMENT{update new bases for each layer by performing SVD.}}
            \STATE$k \gets$ criteria$(\hat{R}^{l},R^{l}, \epsilon_{th})$ \textcolor{blue}{\COMMENT{see Eq. \eqref{eq14}.}}
            \STATE$\mathbf{M}_{l} \gets \left[\mathbf{M}_{l},U^{l}\left[0:k\right]\right]$
        \ENDFOR
        \RETURN $\mathbf{M}$
    \end{algorithmic} 
\end{algorithm}

\subsection{Pseudo-code for RCL}
\label{rcl}
As shown in Alg. \ref{algall}, we present the overall algorithm for RCL.
\begin{algorithm}
    \renewcommand{\algorithmicrequire}{\textbf{Input:}}
	\caption{Algorithm of RCL} 
	\label{algall} 
	\begin{algorithmic}[1]
		\REQUIRE learning rate $\eta$, loss function $\mathcal{L}$, weight value range $\phi$,
        mixup proportion$\lambda$, unif and align proportion $\kappa$, radius $\rho$
		\STATE Initialize: orthogonal bases memory $\{\mathbf{M}_{l,t}\}_{l=1}^{L}: \mathbf{M}_{l,1} \gets \left[\ \right]$, $\mathbf{W} \gets \mathbf{W}_{0}$
		\FOR{$task\ t=1,2,\dots,T$}
            \IF{$t=1$}
            \STATE $\Tilde{\theta}\gets \theta+\epsilon \cdot \phi, \epsilon \sim \mathcal{N}(0,1),\theta \gets \mathbf{W}_{0}$ \textcolor{blue}{\COMMENT{reparameterization, $\phi$ is a random disturbance.}}
            \FOR{$ep\ =1,2,\dots,epochs$}
                \FOR{$l\ =1,2,\dots,L$}
                    \STATE $\hat{\phi_{l}} \gets \phi_{l}-\eta \cdot \nabla_{\phi}\mathcal{L}_{\mathrm{unif}}^{t}(\theta_{l},\Tilde{\theta}_{l})$ \textcolor{blue}{\COMMENT{using uniformity loss to calculate gradient.}}
                    \STATE $\Tilde{\theta}_{l} \gets \theta_{l} + \epsilon \cdot \hat{\phi_{l}}$
                \ENDFOR
            \ENDFOR
            \STATE $\mathbf{W}_{0} \gets \Tilde{\theta}$ \textcolor{blue}{\COMMENT{update initialization parameters.}}
            \ENDIF
            \FOR{$ep\ =1,2,\dots,epochs$}
                \FOR{$batch\ B_{n} \sim \mathcal{D}^{t}$}
                    \STATE $B_{n}^{'} \gets (\Tilde{X},\Tilde{Y}) \gets$ Mixup$(B_{n},\gamma),\gamma \gets Beta(\alpha,\alpha)$
                    \STATE $\hat{\varepsilon} \gets$ using Eq. \eqref{var} \textcolor{blue}{\COMMENT{worst-case data perturbation.}}
                    \STATE $\hat{\upsilon} \gets$ using Eq. \eqref{ups} \textcolor{blue}{\COMMENT{worst-case weight perturbation.}}
                    \STATE $\hat{\gamma} \gets$ Clamp$(\gamma+\hat{\varepsilon},0,1),\hat{\mathbf{W}} \gets \mathbf{W}+\hat{\upsilon}$ \textcolor{blue}{\COMMENT{control the data perturbation within [0,1].}}
                    \STATE $B_{n}^{''} \gets (\Tilde{X},\Tilde{Y}) \gets$ Mixup$(B_{n},\hat{\gamma})$
                    \STATE $g^{t} \gets \nabla_{\mathbf{W}}(\mathcal{L}^{t}(B_{n})+\lambda \mathcal{L}^{t}(B_{n}^{''})+\kappa \mathcal{L}_\mathrm{UA}^{t}(B_{n},B_{n}^{''}))\Big|_{\hat{\mathbf{W}}}$ \textcolor{blue}{\COMMENT{get robust gradient.}}
                    \STATE $\mathbf{W}^l=\mathbf{W}^l-\eta \cdot \hat{g}^{t,l}$ \textcolor{blue}{\COMMENT{update parameters using gradient after projection.}}
                \ENDFOR
            \ENDFOR 
            \STATE $\mathbf{M} \gets $updateGPM$(\mathbf{M},f,\epsilon_{th},\mathcal{D}_{train}^{t})$
        \ENDFOR
	\end{algorithmic} 
\end{algorithm}

\section{Experimental Details}
\label{ex}
\subsection{Datasets}
We evaluate our method on CIFAR-100, 5-Datasets and MiniImageNet datasets. Tab. \ref{d} summarizes the statistics of four datasets used in our experiments.
\begin{table}
    \caption{Dataset Statistics.}
    \label{d}
    \centering
    \renewcommand{\arraystretch}{1.3}
    \scalebox{0.8}{
    \begin{tabular}{cc|ccccc}
    \toprule
    \multicolumn{2}{c}{Datasets}&\#Tasks&\#Classes&\#Train&\#Valid&\#Test\\
    \hline
    CIFAR-100&&10&10&4,750&250&1,000\\
    \hline
    MiniImageNet&&20&5&2,450&50&500\\
    \hline
    \multirow{5}{*}{5-Datasets}&CIFAR-10&\multirow{5}{*}{5}&10&47500&2500&10000\\
    &MNIST&&10&57000&3000&10000\\
    &SVHN&&10&69595&3662&26032\\
    &Fashion MNIST&&10&57000&3000&10000\\
    &NotMNIST&&10&16011&842&1873\\
    \bottomrule
    \end{tabular}}
\end{table}

\subsection{Architecture}
Since all tasks are classification tasks, we use cross entropy plus softmax as the prediction probability for all network output layers. In addition, all network layers use Relu as the activation function. The detailed configuration information for each network architecture is described below.

\textbf{AlexNet:} This architecture is the same as GPM. Specifically, the network consists of 3 convolution layers plus 2 linear layers. The number of filters of the convolution layers from bottom to up is 64, 128, and 256 with kernel sizes 4 × 4, 3 × 3, and 2 × 2, respectively. The number of units of two linear layers is 2048. Rectified function is used as activations for all the layers except the classifier layer. After each convolution layer, 2 × 2 max-pooling is applied. Dropout with ratio 0.2 is used for the first two layers and 0.5 for the rest layers.

\textbf{ResNet18:} Following GPM, the 5-Datasets and MiniImageNet datasets used a smaller version of the ResNet18 \cite{7780459} architecture. Specifically, the first seventeen layers are the convolution layer, and the last is a fully connected layer. The convolution kernel size and padding size of all convolution layers are set to 3×3 and 1×1, respectively. The stride of the 1, 6, 10, and 14-th layers is set to 2×2, and the stride of other convolution layers is set to 1×1.

\subsection{Baselines}
Below, we briefly introduce the baseline methods compared in the experiment.
\begin{itemize}
    \item MTL: an oracle baseline that all tasks are learned jointly using the entire dataset at once in a single network. Multitask is not a continual learning strategy but serves as an upper bound on average test accuracy on all tasks.
    \item EWC \cite{kirkpatrick2017overcoming}: a regularization-based method that uses the diagonal of Fisher information to identify important weights.
    \item ER \cite{chaudhry2019tiny}: a simple and competitive method based on reservoir sampling.
    \item A-GEM \cite{chaudhry2018efficient}: a memory-based method that uses the gradient of episodic memory to constrain optimization to prevent forgetting.
    \item OWM \cite{zeng2019continual}: store the input representations of all samples from the old task as subspaces covered by the old task.
    \item TRGP \cite{lin2022trgp}: promoting forward knowledge transfer between tasks through effective analysis of task relevance.
    \item GPM \cite{saha2021gradient}: maintains important gradient subspaces for the old task of orthogonal projection when updating parameters.
    \item FS-DGPM \cite{deng2021flattening}: using adversarial weight perturbation to flatten sharpness.
    \item API \cite{liang2023adaptive}: implement dual gradient projection and determine whether to expand the model based on its plasticity.
    \item DFGP \cite{yang2023data}: improved the method for obtaining weight perturbations based on FS-DGPM and added data perturbations.
\end{itemize}   

\subsection{List of Hyperparameters}
We follow the baseline method’s hyperparameter settings in GPM and we list the hyperparameter configuration for all methods in Tab. \ref{hy}.
\begin{table}
    \caption{List of hyperparameters for the baselines and our approach. The ‘lr’ denotes the initial learning rate. We represent 10-Split CIFAR-100 as ‘CIFAR’, 5-Datasets as ‘FIVE’ and Split miniImageNet as ‘MIIMG’.}
    \resizebox{\linewidth}{!}{
    \label{hy}
    \centering
    \renewcommand{\arraystretch}{1.3}
    \begin{tabular}{ll}
    \toprule
    Methods&Hyperparameters\\
    \hline
    MTL&lr : 0.05 (CIFAR), 0.1 (FIVE,MIIMG)\\
    \hline
    EWC&lr : 0.03 (FIVE, MIIMG), 0.05 (CIFAR); regularization coefficient : 5000 (CIFAR, FIVE, MIIMG)\\
    ER&lr : 0.05 (CIFAR), 0.1 (FIVE, MIIMG); memory size: 2000 (CIFAR), 500 (MIIMG), 3000 (FIVE)\\
    A-GEM&lr : 0.05 (CIFAR), 0.1 (FIVE, MIIMG); memory size: 2000 (CIFAR), 3000 (FIVE), 500 (MIIMG)\\
    OWM&lr : 0.01 (CIFAR)\\
    TRGP&lr: 0.01 (CIFAR), 0.1 (FIVE, MIIMG) K: 2 (CIFAR, FIVE, MIIMG) $\epsilon$: 0.5 (CIFAR, FIVE, MIIMG)\\
    GPM&lr : 0.01 (CIFAR), 0.1 (FIVE, MIIMG); representation size : 100 (FIVE, MIIMG), 125 (CIFAR)\\
    FS-DGPM&lr : 0.01 (CIFAR), 0.1 (MIIMG); memory size: 1000 (CIFAR, MIIMG)\\
    API&lr: 0.01 (CIFAR), 0.1 (FIVE, MIIMG) K: 10 (CIFAR, FIVE, MIIMG) $\rho$: 0.5 (CIFAR, FIVE, MIIMG)\\
    DFGP&\makecell[l]{lr : 0.01 (CIFAR), 0.1 (FIVE, MIIMG); $\alpha$: 20 (CIFAR, FIVE, MIIMG); $\lambda$: 0.1 (CIFAR), 0.001 (FIVE),\\0.01 (MIIMG); $\rho$: 0.05 (CIFAR, FIVE, MIIMG)}\\
    \hline
    RCL&\makecell[l]{lr : 0.01 (CIFAR), 0.1 (FIVE, MIIMG); $\alpha$: 20 (CIFAR, FIVE, MIIMG); $\lambda$: 0.1 (CIFAR), 1e-04 (FIVE),\\0.1 (MIIMG); $\kappa$: 1 (CIFAR), 1e-05 (FIVE), 0.1 (MIIMG); $\phi$: 1e-04 (CIFAR), 1e-05 (FIVE), 1e-11 (MIIMG)}\\
    \bottomrule
    \end{tabular}
    }
\end{table}
For our method, we search for $\lambda$ for the loss on perturbed data and $\kappa$ for the loss on alignment and uniformity in [0.001, 0.01, 0.1]. We also search for $\phi$ for weight robustness value range in [1e-12, 1e-04]. For each method, we run five random seeds and report the mean and standard deviation.

\subsection{Training Details}
We follow the training settings of GPM \cite{saha2021gradient}, including training epochs and batch size for each dataset. For CIFAR-100, we trained 200 epochs per task and set the batch size to 64. For 5-Datasets and MiniImageNet, we trained 100 epochs per task and set the batch size to 64. All datasets use SGD as the base optimizer. All experiments were run on RTX 4090 (24GB). 

\section{Additional Experimental Results}
\label{adex}

\textbf{A discussion of hyperparameter $\phi$.} $\phi$ is a hyperparameter representing the range of parameter values trained during robust pretraining. As shown in Tab. \ref{wr}, we tested the impact of different values of $\phi$ on accuracy on CIFAR-100, and it can be seen that different ranges of $\phi$ have varying degrees of impact on accuracy. Overall, their accuracy exceeds the optimal value of the DFGP method. Even if $\phi$ takes a very small value (e.g. 1e-08), its accuracy is still better than DFGP (i.e., 75.26). However, there is no clear positive correlation between accuracy and $\phi$, and the optimal range of values varies for different neural network models. For the AlexNet architecture used in the CIFAR-100 dataset, the optimal value of $\phi$ is 1e-04. 
\begin{table}
    \caption{Sensitivity analysis of weight range $\phi$ on CIFAR-100.}
    \label{wr}
    \centering
    \renewcommand{\arraystretch}{1.3}
    \begin{tabular}{cccccc}
    \toprule
    $\phi$&1e-04&1e-05&1e-06&1e-07&1e-08\\
    \hline
    ACC(\%)&75.43$\pm$0.09&75.17$\pm$0.10&75.21$\pm$0.08&75.56$\pm$0.06&75.26$\pm$0.04\\
    \bottomrule
    \end{tabular}
\end{table}

\textbf{A discussion of hyperparameter $\kappa$.} $\kappa$ is a hyperparameter that represents the proportion of alignment and uniformity loss in overall loss optimization. As shown in Tab. \ref{ua}, we tested the accuracy under different $\kappa$ values in CIFAR-100 and found that for smaller values (e.g. 0.6), the improvement in accuracy was also relatively small (i.e., 74.57). But when $\kappa$ equals 0.9, the accuracy reaches 75.50, which is 0.07 higher than when $\kappa$ equals 1. When the value of $\kappa$ is 0, the overall loss mathematical expression of the model is consistent with DFGP.
\begin{table}
    \caption{Sensitivity analysis of the weights of alignment and uniformity loss $\kappa$ on CIFAR-100.}
    \label{ua}
    \centering
    \renewcommand{\arraystretch}{1.3}
    \begin{tabular}{cccccc}
    \toprule
    $\kappa$&1&0.9&0.8&0.7&0.6\\
    \hline
    ACC(\%)&75.43$\pm$0.09&75.50$\pm$0.10&75.35$\pm$0.22&75.06$\pm$0.03&74.57$\pm$0.30\\
    \bottomrule
    \end{tabular}
\end{table}

\textbf{A discussion of random seed.} Conducting experiments on the selection of random seeds can eliminate the influence of random factors on the model. Therefore, we randomly selected 5 random seeds to measure their average accuracy and calculated the standard deviation. The smaller the standard deviation, the more stable the method is. We will compare RCL with DFGP. As shown in Tab. \ref{seed}, the standard deviation of DFGP is 0.303, while the standard deviation of RCL is only 0.09. This indicates that our method is more stable and not an illusion of performance improvement due to random factors.
\begin{table}[H]
    \caption{The averaged accuracy and standard deviation $\sigma$ of different random seed.}
    \label{seed}
    \centering
    \renewcommand{\arraystretch}{1.3}
    \begin{tabular}{cccccc|c}
    \toprule
    Random Seed&1&2&10&50&100&$\sigma$\\
    \hline
    DFGP&74.02&74.22&74.41&74.89&74.15&0.303\\
    RCL&75.42&75.44&75.39&75.46&75.21&0.09\\
    \bottomrule
    \end{tabular}
\end{table}

\textbf{A discussion of training time.}
We provide a time-to-convergence comparison of FS-DGPM, DFGP and RCL on CIFAR-100. The experimental results are shown in Tab. \ref{ti}, FS-DGPM is the least efficient because it iteratively needs to perform iterative adversarial weight perturbations on new task data and old task data in memory and needs to update the scale of the base. DFGP calculates weight perturbations and data perturbations through direct differentiation, greatly reducing convergence time. Our method RCL, on the basis of DFGP, adds training $\phi$ and additional feature distribution loss calculation. Even so, the convergence time does not increase significantly, indicating that our method does not increase the computational complexity of the model. 
\begin{table}[H]
    \caption{Analysis of training time on CIFAR-100.}
    \label{time}
    \centering
    \renewcommand{\arraystretch}{1.3}
    \begin{tabular}{c|ccc}
    \toprule
    Methods&FS-DGPM&DFGP&RCL(ours)\\
    \hline
    Time Cost (min)&148.37&23.81&24.42\\
    \bottomrule
    \end{tabular}
    \label{ti}
\end{table}

\end{document}